\documentclass[11pt,a4paper]{article}
\usepackage[hyperref]{emnlp-ijcnlp-2019}
\usepackage{times}
\usepackage{latexsym}

\usepackage{graphicx}
\usepackage{multirow}
\usepackage{booktabs}
\usepackage{todonotes}

\usepackage{url}

\usepackage{bm}
\usepackage{amsfonts, amsmath}
\usepackage{amsthm}
\usepackage{bbm}
\usepackage{url}
\usepackage{arydshln}
\usepackage{tabularx}

\usepackage{color}
\definecolor{blue}{RGB}{0, 93, 170}			%Go Big Blue!
\definecolor{darkgreen}{RGB}{0, 102, 0}

\aclfinalcopy % Uncomment this line for the final submission

\newcommand{\e}[1]{{\small $#1$}}

\DeclareMathOperator*{\argmax}{\arg\!\max}

\newtheorem{theorem}{Theorem}

\newtheorem{lemma}{Lemma}[theorem]

\makeatletter
\def\adl@drawiv#1#2#3{%
        \hskip.5\tabcolsep
        \xleaders#3{#2.5\@tempdimb #1{1}#2.5\@tempdimb}%
                #2\z@ plus1fil minus1fil\relax
        \hskip.5\tabcolsep}
\newcommand{\cdashlinelr}[1]{%
  \noalign{\vskip\aboverulesep
           \global\let\@dashdrawstore\adl@draw
           \global\let\adl@draw\adl@drawiv}
  \cdashline{#1}
  \noalign{\global\let\adl@draw\@dashdrawstore
           \vskip\belowrulesep}}
\makeatother

\title{Rethinking Cooperative Rationalization: Introspective Extraction and Complement Control}

\author{Mo Yu\thanks{\, Authors contributed equally to this work.}$~~^\heartsuit$  \qquad
  Shiyu Chang$^{*~\clubsuit \heartsuit}$ \qquad Yang Zhang$^{*~\clubsuit \heartsuit}$ \qquad
  Tommi S. Jaakkola$~^\spadesuit$ \\
  $^\heartsuit$ IBM Research \quad $^\clubsuit$ MIT-IBM Watson AI Lab \quad $^\spadesuit$ MIT \\
  {\small
  \texttt{yum@us.ibm.com} \quad \texttt{\{shiyu.chang, yang.zhang2\}@ibm.com} \quad \texttt{tommi@csail.mit.edu}
  }
  }

\date{}

\begin{document}
\maketitle

%!TEX root = emnlp-ijcnlp-2019.tex

\begin{abstract}

Selective rationalization has become a common mechanism to ensure that predictive models reveal how they use any available features. The selection may be soft or hard, and identifies a subset of input features relevant for prediction. The setup can be viewed as a co-operate game between the selector (aka rationale generator) and the predictor making use of only the selected features. The co-operative setting may, however, be compromised for two reasons. First, the generator typically has no direct access to the outcome it aims to justify, resulting in poor performance. Second, there's typically no control exerted on the information left outside the selection. We revise the overall co-operative framework to address these challenges. We introduce an introspective model which explicitly predicts and incorporates the outcome into the selection process. Moreover, we explicitly control the rationale complement via an adversary so as not to leave any useful information out of the selection. We show that the two complementary mechanisms maintain both high predictive accuracy and lead to comprehensive rationales.\footnote{The code and data for our method is publicly available at \url{https://github.com/Gorov/three_player_for_emnlp}.}

\end{abstract}

\section{Introduction}
\label{sec:intro}

%!TEX root = emnlp-ijcnlp-2019.tex

Rapidly expanding applications of complex neural models also bring forth criteria other than mere performance. For example, medical   \cite{yala2019deep} and other high-value decision applications require some means of verifying reasons for the predicted outcomes. This area of self-explaining models in the context of NLP applications has primarily evolved along two parallel tracks. On one hand, we can design neural architectures that expose more intricate mechanisms of reasoning such as module networks \cite{andreas2016learning,andreas2016neural,johnson2017inferring}. While important, such approaches may still require adopting specialized designs and architectural choices that do not yet reach accuracies comparable to black-box approaches. On the other hand, we can impose limited architectural constraints in the form of selective rationalization~\cite{lei2016rationalizing,li2016understanding,chen2018learning,chen2018shapley} where the goal is to only expose the portion of the text relevant for prediction. The selection is done by a separately trained model called rationale generator. The resulting text selection can be subsequently used as an input to an unconstrained, complex predictor, \emph{i.e.}, architectures used in the absence of any rationalization.\footnote{{Therefore the selective rationalization approach is easier to be adapted to new applications, such as question answering~\cite{yu2018learning}, image classification~\cite{chen2018learning} and medical image analysis~\cite{yala2019deep}.}} The main challenge of this track is how to properly coordinating the rationale generator with the powerful predictor operating on the selected information during training.

\begin{table}[t]
\small
\centering
\begin{tabular}{ll}
\toprule
\textbf{Label:} negative \\
\multicolumn{2}{p{7cm}}{\textbf{Original Text:} \textit{really \textcolor{red}{\textit{cloudy}} , \textcolor{red}{\textit{lots of sediment}} , washed out yellow color . looks \textcolor{red}{\textit{pretty gross}} , actually , like swamp water . \textcolor{red}{\textit{no head}} , \textcolor{red}{\textit{no lacing}} .}} \\
% \midrule
\cdashlinelr{0-1}
\textbf{Rationale from \cite{lei2016rationalizing}}:\\ \multicolumn{2}{p{7cm}}{[\textit{``really cloudy lots'', ``yellow'', ``no'', ``no''}]} \\
\textbf{Rationale from cooperative introspection model}: \\
\multicolumn{2}{p{7cm}}{[\textit{``. looks'', ``no'', ``no''}]}\\
\textbf{Rationale from our full model}: \\
\multicolumn{2}{p{7cm}}{[\textit{``cloudy'', ``lots'', ``pretty gross'', ``no lacing''}]}\\
\bottomrule
\end{tabular}
\caption{\small An example of the rationales extracted by different models on the sentiment analysis task of beer reviews (appearance aspect).  \textcolor{red}{\emph{Red}} words are human-labeled rationales.  Details of the experiments can be found in Appendix \ref{app:taos_degeneration}.}
\label{tab:example}
\vspace{-0.1in}
\end{table}

In this paper, we build on and extend selective rationalization. The selection process can be thought of as a cooperative game between the generator and the predictor operating on the selected, partial input text. The two players aim for the shared goal of achieving high predictive accuracy, just having to operate within the confines imposed by rationale selection (a small, concise portion of input text). The rationales are learned entirely in an unsupervised manner, without any guidance other than their size, form. An example of ground-truth and learned rationales are given in Table \ref{tab:example}.
 
The key motivation for our work arises from the potential failures of cooperative selection. Since the generator typically has no direct access to the outcome it aims to justify, the learning process may converge to a poorly performing solution. Moreover, since only the selected portion is evaluated for its information value (via the predictor), there is typically no explicit control over the remaining portion of the text left outside the rationale. These two challenges are complementary and should be addressed jointly. 

% \paragraph{Performance} 
\noindent \textbf{Performance } 
The clues in text classification tasks are typically short phrases~\cite{zaidan2007using}. However, diverse textual inputs offer a sea of such clues that may be difficult to disentangle in a manner that generalizes to evaluation data. Indeed, the generator may fail to disentangle the information about the correct label, offering misleading rationales instead. Moreover, as confirmed by the experiments presented in this paper, the collaborative nature of the game may enable the players to select a sub-optimal communication code that does not generalize, rather overfits the training data. Regression tasks considered in prior work typically offer greater feedback for the generator, making it less likely that such communication patterns would arise.  

We address these concerns by proposing an \emph{introspective rationale generator}. The key idea is to force the generator to explicitly understand what to generate rationales for. Specifically, we make the label that would be predicted with the full text as an additional input to the generator thereby ensuring better overall performance. 

% \paragraph{Rationale quality} 
\noindent \textbf{Rationale quality } 
The cooperative game setup does not explicitly control the information left out of the rationale. As a result, it is possible for the rationales to degenerate as in containing only select words without the appropriate context. In fact, the introspective generator proposed above can aggravate this problem. With access to the predicted label as input, the generator and the predictor can find a communication scheme by encoding the predicted label with special word patterns (\emph{e.g.} highlighting ``.'' for positive examples and ``,'' negative ones). Table \ref{tab:example} shows such degenerate cases for the two cooperative methods.

In order to prevent degenerate rationales, we propose a \emph{three-player game} that renders explicit control over the unselected parts. In addition to the generator and the predictor as in conventional cooperative rationale selection schemes, we add a third adversarial player, called the \emph{complement predictor}, to regularize the cooperative communication between the generator and the predictor. The goal of the complement predictor is to predict the correct label using only words left out of the rationale. During training, the generator aims to fool the complement predictor while still maintaining high accuracy for the predictor. This ensures that the selected rationale must contain all/most of the information about the target label, leaving out irrelevant parts, within size constraints imposed on the rationales. 

We also theoretically show that the equilibrium of the three-player game guarantees good properties for the extracted rationales. Moreover, we empirically show that (1) the three-player framework on its own helps cooperative games such as \cite{lei2016rationalizing} to improve both predictive accuracy and rationale quality; (2) by combining the two solutions -- introspective generator and the three player game -- we can achieve high predictive accuracy and non-degenerate rationales.

\section{Problem Formulation}
This section formally defines the problem of rationalization, and then proposes a set of conditions that desirable rationales should satisfy, which addresses problems of previous cooperative frameworks.  Here are some notations throughout this paper. Bolded upper-cased letters, \emph{e.g.} \e{\bm X}, denote random vectors; unbolded upper-cased letters, \emph{e.g.} \e{X}, denote random scalar variables; bolded lower-cased letters, \emph{e.g.} \e{\bm x}, denote deterministic vectors or vector functions; unbolded lower-cased letters, \emph{e.g.} $x$, denote deterministic scalars or scalar functions. \e{p_X(\cdot | Y)} denotes conditional probability density/mass function  conditional on \e{Y}. \e{H(\cdot)} denotes Shannon entropy. \e{\mathbb{E}[\cdot]} denotes expectation.

% ----------------------------------
\subsection{Problem Formulation of Rationalization}

The target application here is text classification on data tokens in the form of \e{\{(\bm X, Y)\}}. Denote \e{\bm X = \bm X_{1:L}} as a sequence of words in an input text with length \e{L}. Denote \e{Y} as a label.  Our goal is to generate a rationale, denoted as \e{\bm r(\bm X) = \bm r_{1:L}(\bm X)}, which is a selection of words in \e{\bm X} that accounts for \e{Y}. Formally, \e{\bm r(\bm X)} is a hard-masked version of \e{\bm X} that takes the following form at each position \e{i}:
\begin{equation}
\small
    \bm r_i(\bm X) = \bm z_i(\bm X) \cdot \bm X_i,
    \label{eq:rat}
\end{equation}
where \e{\bm z_i \in \{0, 1\}^N} is the binary mask. Many previous works \cite{lei2016rationalizing,chen2018learning} follows the above definition of rationales.  In this work, we further define the complement of rationale, denoted as \e{\bm r^c(\bm X)}, as
\begin{equation}
\small
    \bm r_i^c(\bm X) = (1 - \bm z_i(\bm X)) \cdot \bm X_i.
    \label{eq:ratc}
\end{equation}
For notational ease, define
\begin{equation}
    \small
    \bm R = \bm r(\bm X),\quad \bm R^c = \bm r^c(\bm X),\quad \bm Z = \bm z(\bm X).
    \label{eq:notation}
\end{equation}

% ----------------------------------
\subsection{Rationale Conditions}
\label{ssec:axiomatic}

An ideal rationale should satisfy the following conditions.

\noindent \textbf{Sufficiency:} \e{\bm R} is sufficient to predict \e{Y}, \emph{i.e.}
\begin{equation}
\small
    p_Y(\cdot | \bm R) = p_Y(\cdot | \bm X).
    \label{eq:suff}
\end{equation}

\noindent \textbf{Comprehensiveness:} \e{\bm R^c} does not contain sufficient information to predict \e{Y}, \emph{i.e.}
\begin{equation}
\small
    H(Y| \bm R^c) \geq H(Y | \bm R) + h,
    \label{eq:nondegen}
\end{equation}
for some constant $h$.

\noindent \textbf{Compactness:} the segments in \e{\bm X} that are included in \e{\bm R} should be sparse and consecutive, \emph{i.e.,}
\begin{equation}
\small
    \sum_i \bm Z_i \leq s, \quad \sum_i |\bm Z_i - \bm Z_{i-1}| \leq c,
    \label{eq:compact}
\end{equation}
for some constants $s$ and $c$.

Here is an explanation of what each of these conditions means. The sufficiency condition is the core one of a legitimate rationale, which essentially stipulates that the rationale maintains the same predictive power as \e{\bm X} to predict \e{Y}. The compactness condition stipulates that the rationale should be continuous and should not contain more words than necessary. For example, without the compactness condition, a trivial solution to Eq.~\eqref{eq:suff} would be \e{\bm X} itself. The first inequality in Eq.~\eqref{eq:compact} constrains the sparsity of rationale, and the second one constrains the continuity.  The comprehensiveness condition requires some elaboration, which we will discuss in the next subsection.

% ----------------------------------
\subsection{Comprehensiveness and Degeneration}

There are two justifications of the comprehensiveness condition.  First, it regulates the information outside the rationale, so that the rationale contains all the relevant and useful information, hence the name comprehensiveness.  Second and more importantly, we will show there exists a failure case, called degeneration, which can only be prevented by the comprehensiveness condition.

Degeneration refers to the situation where, rather than finding words in \e{\bm X} that explains \e{Y}, \e{\bm R} attempts to encode the probability of \e{Y} using trivial information, \emph{e.g.} punctuation and position. Consider the following toy example of binary classification (\e{Y \in \{0, 1\}}), where \e{\bm X} can always perfectly predict \e{Y}. 
Then the following rationale satisfies the sufficiency and compactness:
\e{\bm R} includes only the first word of \e{\bm X} when \e{Y=0}, and only the last word when \e{Y=1}. It is obvious that this \e{\bm R} is sufficient to predict \e{\bm Y} (by looking at if the first word or the last word is chosen), and thus satisfies the sufficiency condition. Apparently this \e{\bm R} is perfectly compact (only one word).  However, this rationale does not provide a valid explanation.

Theoretically, any previous cooperative framework may suffer from the above ill-defined problem, if the generator has the potential to accurately guess \e{Y} with sufficient capacity.\footnote{We show such cases of \cite{lei2016rationalizing} in Appendix \ref{app:taos_degeneration}.}
This problem happens because they have no control of the words unselected by \e{\bm R}.
Intuitively, in the presence of degeneration, some key predictors in \e{\bm X} will be left unselected by \e{\bm R}. Thus by looking at the predictive power of \e{\bm R^c}, we can determine if degeneration occurs. Specifically, when degeneration is present, all the useful information is left unselected by \e{\bm R}, and so \e{H(Y | \bm R^c)} is low. That is why the lower bound in Eq.~\eqref{eq:nondegen} rules out the degeneration cases.

\begin{figure*}[ht!]
\centering
  \includegraphics[clip,width=1.5\columnwidth]{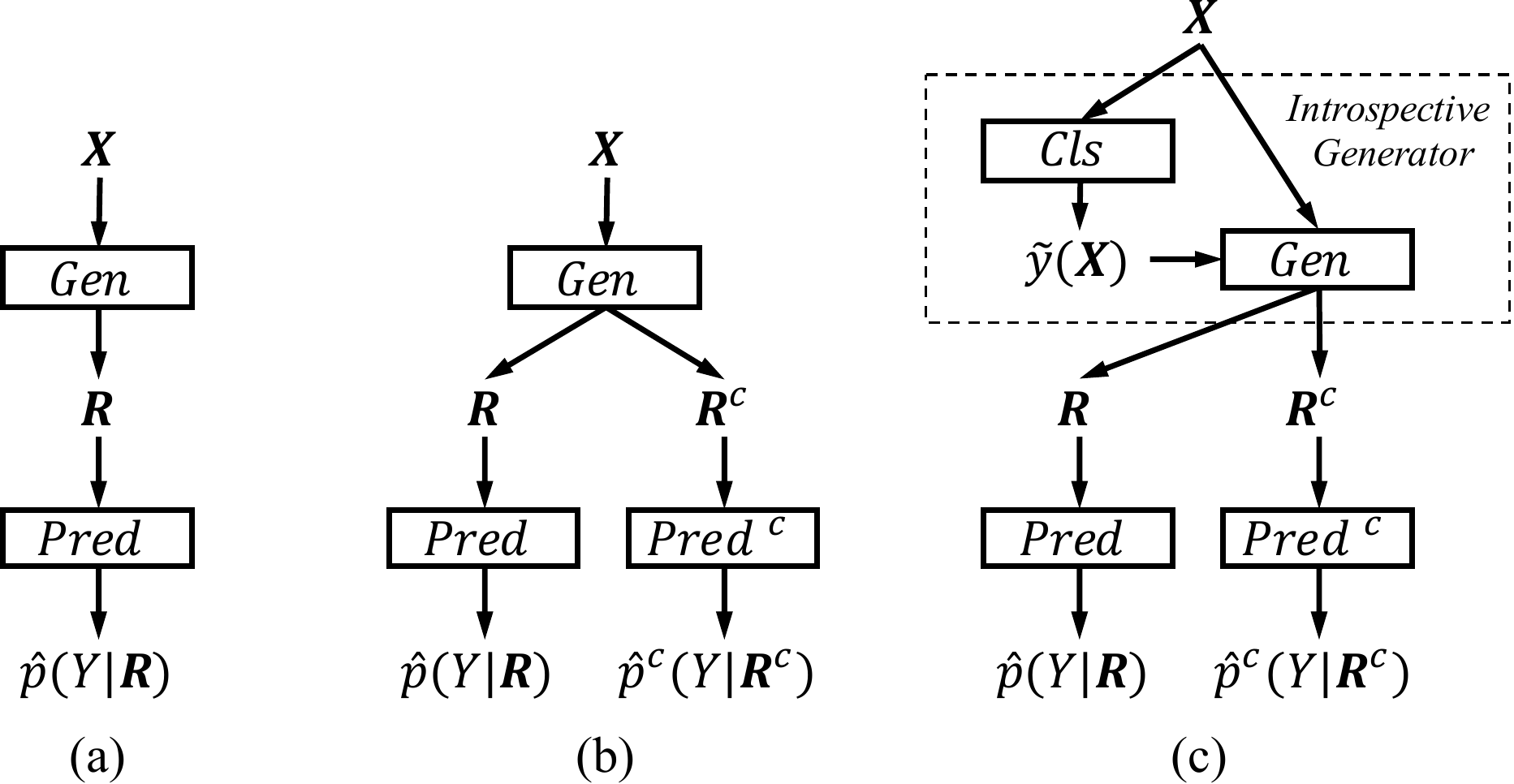}  
  \vspace{-0.1in}
\caption{\small{Illustration of different rationalization frameworks.  (a) The cooperative framework from \cite{lei2016rationalizing}, which consists of two players, \emph{i.e.,} a generator (\emph{Gen}) and a predictor (\emph{Pred}). (b) A straightforward extension of the model (a) with a compliment predictor (\emph{Pred}$^{\,c}$).  The generator plays a cooperative game with the predictor and plays a mini-max game with the compliment predictor.   (c) The introspective three-player framework.  The introspective module first predicts the possible outcome $\tilde{y}$  based on the full texts  $x$ and then generate rationales using both $x$ and $\tilde{y}$.  The third framework (c) is a special case of (b).  Such an inductive bias in model design preserves the predictive performance.  
}}
\label{fig:models}
\end{figure*}

\section{The Proposed Three-Player Models}

This section introduces our new rationalization solutions, which is theoretically guaranteed to be able to find the rationales that satisfy Eqs.~(\ref{eq:suff}-\ref{eq:compact}).

\subsection{The Basic Three-Player Model} 
\label{ssec:three_player_base}

This section introduces our basic three-player model.
The model consists of three players: a \textbf{rationale generator} that generates the rationale \e{\bm R} and its complement \e{\bm R^c} from text, a \textbf{predictor} that predicts the probability of \e{Y} based on \e{\bm R}, a \textbf{complement predictor} that predicts the probability of \e{Y} based on \e{\bm R^c}.

Figure \ref{fig:models}(b) illustrates the basic three-player model.  Compared with \cite{lei2016rationalizing}, as shown in Figure \ref{fig:models}(a), the three-player model introduces an additional complement predictor, which plays a minimax game in addition to the cooperative game in \cite{lei2016rationalizing}. For clarity, we will describe the game backward, starting from the two predictors followed by the generator. 

\noindent \textbf{Predictors:} The predictor estimates probability of \e{Y} conditioned on \e{\bm R}, denoted as \e{\hat{p}(Y|\bm R)}. The complement predictor estimates probability of \e{Y} conditioned on \e{\bm R^c}, denoted as \e{\hat{p}^c(Y|\bm R)}. Both predictors are trained using the cross entropy loss, \emph{i.e.}
\begin{equation}
    \small
    \begin{aligned}
    & \mathcal{L}_p = \min_{\hat{p}(\cdot, \cdot)} -H(p(Y | \bm R); \hat{p}(Y | \bm R)) \\
    & \mathcal{L}_c = \min_{\hat{p}^c(\cdot, \cdot)} -H(p(Y | \bm R^c); \hat{p}^c(Y | \bm R^c)),
    \end{aligned}
    \label{eq:loss_pred}
\end{equation}
where \e{H(p;q)} denotes the cross entropy between \e{p} and \e{q}. \e{p(\cdot | \cdot)} denotes the empirical distribution. It is worth emphasizing that \e{\mathcal{L}_p} and \e{\mathcal{L}_c} are both functions of the generator.

\noindent \textbf{Generator:} The generator extracts \e{\bm R} and \e{\bm R^c} by generating the rationale mask, \e{\bm z(\cdot)}, as shown in Eqs.~(\ref{eq:rat}-\ref{eq:ratc}). Specifically, \e{\bm z(\cdot)} is determined by minimizing the weighted combination of four losses:
\begin{equation}
\small
    \min_{\bm z(\cdot)} \mathcal{L}_p + \lambda_g \mathcal{L}_g + \lambda_s \mathcal{L}_s + \lambda_c \mathcal{L}_c,
    \label{eq:loss_gen}
\end{equation}
where \e{\mathcal{L}_g} encourages the gap between \e{\mathcal{L}_p} and \e{\mathcal{L}_c} to be large, \emph{i.e.}
\begin{equation}
\small
    \mathcal{L}_g = \max \{\mathcal{L}_p - \mathcal{L}_c + h, 0\}.
    \label{eq:Lg}
\end{equation}
It stipulates the comprehensiveness property of the rationale (Eq.~\eqref{eq:nondegen}). Intuitively, if the complement rationale is less informative of \e{Y} than the rationale, then \e{\mathcal{L}_c} should be larger than \e{\mathcal{L}_p}.

\e{\mathcal{L}_s} and \e{\mathcal{L}_c} impose the sparsity and continuity respectively, which correspond to Eq.~\eqref{eq:compact}:
\begin{equation}
    \small
    \begin{aligned}
    &\mathcal{L}_s = \max\Big\{\sum_i \bm Z_i - s, 0\Big\},\\ &\mathcal{L}_c = \sum_i \max\left\{||\bm Z_i - \bm Z_{i-1}|| - c, 0\right\}.
    \end{aligned}
    \label{eq:Ls_Lc}
\end{equation}

From Eq.~\eqref{eq:loss_pred}, we can see that the generator plays a \emph{cooperative game} with the predictor, because both tries to maximize the predictive performance of \e{\bm R}.  On the other hand, the generator plays an \emph{adversarial game} with the complement predictor, because the latter tries to maximize the predictive performance of \e{\bm R^c}, but the former tries to reduce it. Without the complement predictor, and thus the loss \e{\mathcal{L}_g}, the framework reduces to the method in \cite{lei2016rationalizing}.

\paragraph{Training} During training, the three players perform gradient descent steps with respect to their own losses. 
For the generator, Since $\bm z(\bm X)$ is a set of binary variables, we cannot apply the regular gradient descent algorithm. Instead we will use policy gradient~\cite{Williams1992} to optimize the models. 
We maximize the reward that is defined as the negative loss in Eq. \eqref{eq:loss_gen}.
In order to have bounded rewards for training stability, the negative losses \e{\mathcal{L}_p} and \e{\mathcal{L}_c} are replaced with accuracy.
% For the predictors, the reward is defined by accuracy, which is a proxy for the negative losses in Eq.~\eqref{eq:loss_pred}.

\paragraph{Theoretical guarantees}
The proposed framework is able to obtain a rationale that simultaneously satisfies the conditions in Eqs.~\eqref{eq:suff} to \eqref{eq:compact}, as stated in the following theorem:
% In this section, we will show that how the proposed framework is able to obtain a rationale that simultaneously satisfy the conditions in Eqs.~\eqref{eq:suff} to \eqref{eq:compact}, as stated in the following theorem.
\begin{theorem}
A rationalization scheme \e{\bm z(\bm X)} that simultaneously satisfies Eqs.~\eqref{eq:suff}-\eqref{eq:compact} is the global optimizer of Eq.~\eqref{eq:loss_gen}.
\label{thm:main}
\end{theorem}
The proof is given in Appendix~\ref{app:theorem}. The basic idea is that there is a correspondence between each term in Eq.~\eqref{eq:loss_gen} and each of the properties Eqs.~\eqref{eq:suff}-\eqref{eq:compact}. The minimization of each loss term is equivalent to satisfying the corresponding property.

\subsection{The Introspection Generator}

As discussed in Section \ref{sec:intro}, in the existing generator-predictor framework, the generator may fail to disentangle the information about the correct label, offering misleading rationales instead. To address this problem, we propose a new generator module, called the \textbf{Introspective Generator}, which explicitly predicts the label before making rationale selections.

Figure \ref{fig:models}(c) illustrates the model with the introspection generator.  Specifically, the improved generator still fits into the basic three-player framework in Section \ref{ssec:three_player_base}. The only difference is in how the generator generates the mask \e{\bm z(\bm X)}, which now breaks down into two steps.

First, the module uses a \emph{regular classifier} that takes the input \e{\bm X} and predicts the label, denoted \e{\tilde{y}(\bm X)}.  For classification tasks, we use the maximum likelihood estimate, \emph{i.e.}
\begin{equation}
    \small
    % \hat{y}(\bm X) = \argmax_y ~\hat{p}(Y=y | \bm X)
    \tilde{y}(\bm X) = \argmax_y ~\tilde{p}(Y=y | \bm X),
\end{equation}
where \e{\tilde{p}(Y=y | \bm X)} is the predicted probability by maximizing the cross entropy, which is pretrained. 

Second, a label-aware rationale generator generates the binary mask of the rationales, \emph{i.e.}
\begin{equation}
    \small
    z(\bm X) = \tilde{z}(\bm X, \tilde{y}(\bm X)).
\end{equation}
Note that \e{\tilde{y}} is a function of \e{\bm X}, so the entire introspective generator is essentially a function of \e{\bm X}. In this way, all the formulations in Section \ref{ssec:three_player_base} and the Theorem \ref{thm:main} still hold for the three-player game with the introspective generator.

In the implementation, the classifier can use the same architecture like the predictor and the complement predictor. The generator is of the same architecture in Section \ref{ssec:three_player_base}, but with the additional input of \e{\hat{y}(\bm X)}.

\paragraph{Remark on degeneration} Obviously, when working in a cooperative game, the introspection generator will make the degeneration problem more severe: when the classifier \e{\tilde{p}(\cdot | \bm X)} becomes sufficiently accurate during training, the generator only needs to encode the information of \e{\tilde{y}} into \e{\bm R}. 
Therefore our three-player game, while improving over any existing generator-predictor frameworks on its own, is critical for the introspective model.

\begin{table*}[t]
\centering
\small
\begin{tabular}{lcc}
\toprule
\bf Task   & \bf Label & \bf Input Texts \\
% \cmidrule{2-5}  \cmidrule{6-9} 
\midrule
\multirow{2}{*}{Multi-Aspect Beer Review} &  \multicolumn{1}{p{3cm}}{positive (regarding the aspect of \emph{appearance})} & \multicolumn{1}{p{8cm}}{\textit{\textcolor{red}{\textbf{Clear, burnished copper-brown topped by a large beige head that displays impressive persistance and leaves a small to moderate amount of lace in sheets when it eventually departs}}. The nose is sweet and spicy and the flavor is malty sweet, accented nicely by honey and by abundant caramel/toffee notes. There's some spiciness (especially cinnamon), but it's not overdone. The finish contains a moderate amount of bitterness and a trace of alcohol. The mouthfeel is exemplary $\cdots$}}\\ 
%; full and rich, very creamy. Mouthfilling with some mouthcoating as well$\cdots$}} \\
\midrule
Single-Aspect Beer Review & positive & \multicolumn{1}{p{8cm}}{\textit{appearance : dark-brown/black color with a \textcolor{red}{\textbf{huge tan head}} that gradually collapses , leaving \textcolor{red}{\textbf{thick lacing}} .}} \\
\midrule
Relation Classification & Message-Topic(e1,e2) & \multicolumn{1}{p{8cm}}{\textit{It was a friendly \textcolor{red}{\textbf{call$_{e_1}$ to remind}} them \textcolor{red}{\textbf{about the bill$_{e_2}$}} and make sure they have a copy of the invoice}} \\
\bottomrule
\end{tabular}
\caption{Example of the three tasks used for evaluation. The \textcolor{red}{\textbf{\textit{red bold}}} words are sample rationales (created by the authors for illustration purpose only for the last two tasks).}
\label{tab:datasets}
\end{table*}

% --------------------------------------------------------------------------------------
\section{Experimental Settings}

\subsection{Datasets}
\label{ssec:datasets}
We construct three text classification tasks, including two sentiment classification tasks from the BeerAdvocate review dataset \cite{mcauley2012learning}\footnote{http://snap.stanford.edu/data/web-BeerAdvocate.html}, and a more complex relation classification task from SemEval 2010 Task 8~\cite{hendrickx2009semeval}. Table \ref{tab:datasets} gives examples of the above tasks.
Finally, as suggested by an anonymous reviewer, we evaluate on the text matching benchmark AskUbuntu, following~\citet{lei2016rationalizing}. The experimental setting and results are reported in Appendix \ref{app:askubuntu}.

\paragraph{Multi-aspect beer review} 
This is the same data used in~\cite{lei2016rationalizing}. Each review evaluates multiple aspects of a kind of beer, including appearance, smell, palate, and an overall score. For each aspect, a rating $\in$ [0,1] is labeled.  We limit ourselves to the appearance aspect only and use a threshold of 0.6 to create balanced binary classification tasks for each aspect.  Then, the task becomes to predict the appearance aspect of a beer based on multi-aspect text inputs.  The advantage of this dataset is that it enables automatic evaluation of rationale extraction.  The dataset provides sentence-level annotations on about 1,000 reviews, where each sentence is labeled by the aspect it covers.  Note that in this dataset, each aspect is often described by a single sentence with clear polarity.   Thus, a generator can select a sentence based on the topic distribution of words.  The selected sentence often has very high overlap with the ground truth annotations and also contains sufficient information for predicting the sentiment.   This characteristic of the dataset makes it a relatively easy task, and thus we further consider two more challenging tasks.  

\paragraph{Single-aspect beer review} To construct a more challenging task, for each review, we extract the sentences that are specifically about the appearance aspect from the aforementioned BeerAdvocate dataset\footnote{We will release the single-aspect dataset.}, and the task is to predict the sentiment of the appearance only on the extracted sentences.  The details of the dataset construction can be found in Appendix \ref{app:data_construction}.  This new dataset is obviously more challenging in terms of generating meaningful rationales.  This is because the generator is required to select more find-grained rationales.  Since there are no rationale annotations, we rely on subjective evaluations to test the quality of the extracted rationales.

\paragraph{Multi-class relation classification}
To show the generalizability of our proposed approaches to other NLP applications, we further consider the SemEval 2010 Task 8 dataset \cite{hendrickx2009semeval}.  Given two target entities $e_1$ and $e_2$ in a sentence, the goal is to classify the relation type (with directions) between the two entities (including None if there is no relation).   Similar to the single-aspect beer review, it is also a fine-grained rationalization task.  The major difference is that the number of class labels in this task is much larger, and we hope to investigate its effects on degeneration and performance downgrade.

\subsection{Implementation Details}
For both the generators and the two predictors, we use bidirectional LSTMs with hidden dimension 400.  In the introspection generator, the classifier consists of the same bidirectional LSTM, and \e{\bm z(\bm X, \tilde{y})} is implemented as an LSTM sequential labeler with the label $\tilde{y}$ transformed to an embedding vector that serves as the initial hidden states of the LSTM.   For the relation classification task, since the model needs to be aware of the two target entities, we add the relative position features following \citet{nguyen2015combining}.  We map the relative position features to learnable embedding vectors and concatenate them with word embeddings as the inputs to the LSTM encoder of each player.  
All hyper-parameters are tuned on the development sets according to predictive accuracy. In other words, all the models are tuned \textbf{without} seeing any rationale annotations.

% --------------------------------------------------------------------------------------
\section{Experiments}
\label{sec:exp}
% ------------------------------- 
\subsection{Multi-Aspect Beer Review}

\begin{table}[t]
\centering
\small
\begin{tabular}{lcccccc}
\toprule
& \multicolumn{2}{c}{\bf 10\% Highlighting}& \multicolumn{2}{c}{\bf 20\% Highlighting}\\
\bf Model   & \bf Acc & \bf Prec & \bf Acc & \bf Rec \\
\midrule
Random & 65.70 & 17.75 & 72.31 & 19.84\\
\midrule
Lei2016 & 82.05 & 86.14 & 83.88 & 79.98\\
\multirow{1}{*}{ +minimax} & 83.45 & 86.54 & 84.25 & 85.16 \\
\midrule
\multirow{1}{*}{Intros} & 87.10 & 68.37 & 87.50 & 59.63 \\
\multirow{1}{*}{ +minimax} & 86.16 & 85.67 & 86.86 & 79.40 \\
\bottomrule
\end{tabular}
\caption{\small Main results of binary classification on multi-aspect beer reviews.  The \emph{Acc} column shows the accuracy of prediction. The \emph{Prec} and \emph{Rec} are precision and recall on the extracted rationales compared to the human annotations. The accuracy of sentiment prediction with the full texts is \textbf{87.59}.}
\label{tab:beer_results}
\end{table}

Table \ref{tab:beer_results} summarizes the main results on the multi-aspect beer review task.  In this task, a desirable rationale should be both appearance-related and sufficient for sentiment predictions.  Since the average sparsity level of human-annotated rationales is about 18\%, we consider the following evaluation settings.   Specifically, we compare the generated rationales to human annotations by measuring the precision when extracting 10\% of words and the recall for 20\%\footnote{Previous work in \cite{lei2016rationalizing} only evaluates the precision for the 10\% extraction.}.  In addition to the precision/recall, we also report the predictive accuracy of the extracted rationales on predicting the sentiment.

When only 10\% of the words are used, \citet{lei2016rationalizing} has a significant performance downgrade compared to the accuracy when using the whole passage (82.05 v.s. 87.59).  With the additional third player added, the accuracy is slightly improved, which validates that controlling the unselected words improves the robustness of rationales.   On the other hand, our introspection models are able to maintain higher predictive accuracy (86.16 v.s. 82.05) compared to \cite{lei2016rationalizing}, while only sacrificing a little loss on highlighting precision (0.47\% drop).

Similar observations are made when 20\% of the words required to highlight with one exception.  Comparing the model of \cite{lei2016rationalizing} with and without the proposed mini-max module, there is a huge gap of more than 5\% on recall of generated rationales.  This confirms the motivation that the original cooperative game tends to generate less comprehensive rationales, where the three-player framework controls the unselected words to be less informative so the recall is significantly improved.

It is worth mentioning that when a classifier is trained with randomly highlighted rationales (\emph{i.e.} random dropout \cite{srivastava2014dropout} on the inputs), it performs significantly worse on both predictive accuracy and highlighting qualities.  This confirms that extracting concise and sufficient rationales is not a trivial task.   Moreover, our re-implementation of \cite{lei2016rationalizing} for the original regression task achieves 90.1\% precision when highlighting 10\% words, which suggest that rationalization for binary classification is more challenging compared to the regression where finer supervision is available.

In summary, our three-player framework consistently improves the quality of extracted rationales on both of the original \cite{lei2016rationalizing} and the introspective framework.  Particularly, without controlling the unselected words, the introspection model experiences a serious degeneration problem as expected.  With the three-player framework, it manages to maintain both high predictive accuracy and high quality of rationalization. 

% ------------------------------- 

\begin{table}[t]
\centering
\small
\begin{tabular}{lcc||cc}
\toprule
% \midrule
\multirow{2}{*}{\bf Model}   & \multicolumn{2}{c||}{\bf Single-Aspect Beer} & \multicolumn{2}{c}{\bf Relation}  \\
 & \bf Acc & \bf Acc$^{c}$ & \bf Acc & \bf Acc$^{c}$\\
\midrule
All (100\%) & 87.1 & 51.3 & 77.8 & 4.9  \\
Random & 74.3 & 83.1 & 51.8 & 64.9  \\
Lei2016 &  81.4 & 74.6 & 73.5 & 36.8 \\
\multirow{1}{*}{ +minimax} & 82.9 & 73.2 & 75.0 & 30.9\\
\midrule
\multirow{1}{*}{Intros} & 87.0 & 85.8  & 75.2 & 56.6  \\
\multirow{1}{*}{ +minimax} & 85.3 & 73.6 &  76.2 & 29.1 \\
\bottomrule
\end{tabular}
\caption{\small Predictive accuracy on the single-aspect beer appearance review (left) and the relation classification (right).  Acc$^c$ refers to the accuracy of the complement predictor.  We restrict the extracted rationales to be on average eight words and four continuous pieces for the beer review and six words and three pieces for relation classification.  A desired rationalization method will have high Acc and low Acc$^c$. }
\label{tab:sentence_acc}
\end{table}

\subsection{Single-Aspect Beer Review}

In this section, we evaluate the proposed methods on the more challenging single-aspect beer dataset.   Similar to previous experiments, we force all the methods to have comparable highlighting ratio and continuity constraints for fair evaluation.  The highlighting ratio is determined from human estimation on a small set of data.  We report two classification results, which are the accuracy of the predictor and complement predictor.  For both cooperative methods, \emph{i.e.} \cite{lei2016rationalizing} with and without introspection, we train an independent extra predictor on the unselected words from the generator, which does not affect the training of the generator-predictor framework.

From the left part of Table \ref{tab:sentence_acc}, we observe that the original model of \cite{lei2016rationalizing} has a hard time maintaining the accuracy compared to the classifier trained with full texts.  Transforming it into a three-player game helps to improve the performance of the evaluation set while lower the accuracy of the complement predictor.  Since the extracted rationales are in a similar length, these results suggest that learning with a three-player game successfully enforces the generator not to leave informative words unselected.  Similar to the multi-aspect results, the cooperative introspection model suffers from the degeneration problem, which produces a high complement accuracy.  The three-player game yields a more comprehensive rationalization with small losses in accuracy.

% ------------------------------- 
\begin{table}[t]
\centering
\small
\begin{tabular}{lccc}
\toprule
\bf Model  & \bf \%UNK  & \bf Acc   & \bf Acc$_{\textrm{w/o UNK}}$  \\
\midrule
Lei2016 &  43.5 & 63.5 & 69.0  \\
Intros+minimax & \textbf{54.0}$^{\dagger}$ & \textbf{58.0} & \textbf{66.3}\\
\bottomrule
\end{tabular}
\caption{\small Subjective evaluations on the task of controlling the unselected rationale words.  Acc denotes the accuracy in guessing sentiment labels.   Acc$_{\textrm{w/o UNK}}$ denotes the sentiment accuracy for these samples that are not selected as ``UNK'' for the secondary task.  $\dagger$ denotes p-value $<$ 0.005 in t-test.  A desired rationalization method achieves high ``UNK'' rate and performance randomly for the Acc predictions. }
\label{tab:sentence_human_eval}
\end{table}

% ------------------------------- 
\paragraph{Human evaluation}
We further conduct subjective evaluations by comparing the original model of \cite{lei2016rationalizing} with our introspective three-player model.  We mask the extracted rationales and present the unselected words only to the human evaluators.   The evaluators are asked to predict the sentiment label as their first task.  If a rationalizing method successfully includes all informative pieces in the rationale, subjects should have around 50\% of accuracy in guessing the label.   In addition, after providing the sentiment label, subjects are then asked to answer a secondary question, which is whether the provided text spans are sufficient for them to predict the sentiment.  If they believe there are not enough clues and their sentiment classification is based on a random guess, they are instructed to select unknown (denoted as ``UNK'') as the answer to the second question.    Appendix \ref{app:guidelines} elaborates why we design subjective evaluations in such a way in more details.

Table \ref{tab:sentence_human_eval} shows the performance of subjective evaluations.  Looking at the first column of the table, our model is better in confusing human, which gives a higher rate in selecting ``UNK''.   It confirms that the three-player introspective model selects more comprehensive rationales and leave less informative texts unattended.  Furthermore, the results also show that human evaluators offer worse sentiment predictions on the proposed approach, which is also desired and expected.  

% --------------------------------------------------------------
\subsection{Relation Classification}

The predictive performances on the relation classification task are shown in the right part of the Table \ref{tab:sentence_acc}.  We observe consistent results as in previous datasets.  Clearly,  the introspective generator helps the accuracy and the three-player game regularize the complement of the rationale selections.  

% ------------------------------- 
\paragraph{Examples of the extracted rationales}
For relation classification, it is difficult to conduct subjective evaluations because the task requires people to have sufficient knowledge of the schema of relation annotation.   To further demonstrate the quality of generated rationales, we provide some illustrative examples.  Since there is a rich form of supervised signal, \emph{i.e.,} the number of class labels is large, the chance of any visible degeneration of the \citet{lei2016rationalizing}'s model should be low.  However, we still spot quite a few cases.  In the first example, \citet{lei2016rationalizing} fails to highlight the second entity while ours does.  In the second example, the introspective three-player model selects more words than \cite{lei2016rationalizing}.   In this case, the two entities themselves suffice to serve as the rationales.  However, our model preserves the words like ``working''.  This problem might due to the bias of the dataset.  For example, some words that are not relevant to the target entities may still correlate with the labels.   In the case, our model will pick these words as a part of the rationale.

\begin{table}[t]
\small
\centering
\begin{tabular}{ll}
\toprule
\multicolumn{2}{p{7cm}}{\textbf{Original Text}: \textit{{of} the hundreds of strains of avian influenza a \textbf{viruses$_{e_1}$} , only four have caused human \textbf{infections$_{e_2}$} : h5n1 , h7n3 , h7n7 , and h9n2}} \\
\textbf{Label:} \emph{Cause-Effect($e_1$,$e_2$)}\\
\cdashlinelr{0-1}
\multicolumn{2}{p{7cm}}{\textbf{\citet{lei2016rationalizing}}:} \\
\multicolumn{2}{p{7cm}}{[\textit{\textbf{viruses$_{e_1}$} , only four have caused}]} \\
\multicolumn{2}{p{7cm}}{\textbf{Our Intros+minimax}:}\\
\multicolumn{2}{p{7cm}}{[\textit{\textbf{viruses$_{e_1}$}}], [\textit{four have caused human \textbf{infections$_{e_2}$}}]}\\
\midrule \midrule
\multicolumn{2}{p{7cm}}{\textbf{Original Text}: \textit{i spent a year working for a \textbf{software$_{e_1}$ company$_{e_2}$} to pay off my college loans}} \\
\textbf{Label:} \emph{Product-Producer($e_1$,$e_2$)}\\
\cdashlinelr{0-1}
\multicolumn{2}{p{7cm}}{\textbf{\citet{lei2016rationalizing}}:} \\
\multicolumn{2}{p{7cm}}{[\textit{a \textbf{software$_{e_1}$ company$_{e_2}$} to}]} \\
\multicolumn{2}{p{7cm}}{\textbf{Our Intros+minimax}:}\\
\multicolumn{2}{p{7cm}}{[\textit{working for a \textbf{software$_{e_1}$ company$_{e_2}$}}], [\textit{loans}]}\\
\bottomrule
\end{tabular}
\caption{\small Illustrative examples of generated rationales on the relation classification task.   Entities are shown in \textbf{\textit{bold}}.}
\label{tab:qualitative}
\end{table}

% --------------------------------------------------------------------------------------
\section{Related Work}

% ----------------
\paragraph{Model interpretability }
Besides the two major categories of self-explaining models discussed in Section \ref{sec:intro}, model interpretability is widely studied in the general machine learning field.  For example, evaluating feature importance with gradient information \cite{simonyan2013deep,li2016visualizing, sundararajan2017axiomatic} or local perturbations \cite{kononenko2010efficient, lundberg2017unified};  and interpreting deep networks by locally fitting interpretable models~\cite{ribeiro2016should,alvarez2018towards}.  

Besides selective rationalization, the cooperative game has been studied in the latter direction above~\cite{lee2018game}. 
It has also been applied to a relevant problem on summarization~\cite{arumae2019guiding}, where the selected summary should be sufficient for answering questions related to the document. In this problem, the summary is a special type of rationale of a document.
Another related concurrent work~\cite{bastings2019interpretable} proposes differentiable solution to optimize the cooperative rationalization method.

% ----------------
\paragraph{Game-theoretical methods }
Though not having been explored much for self-explaining models, the minimax game setup has been widely used in many machine learning problems, such as self-playing for chess games~\cite{silver2017mastering}, generative models~\cite{goodfellow2014generative} and many tasks that can be formulated as multi-agent reinforcement learning~\cite{busoniu2006multi}.  Our three-player game framework also shares a similar idea with ~\cite{zhang2017aspect,zhao2017learning}, which aim to learn domain-invariant representations with both cooperative and minimax games.

\paragraph{Version 2 update} We recently found an independent effort \cite{carton2018extractive}, which proposed a similar three-player game solution for identifying personal attacks in social media posts. Both \citet{carton2018extractive} and our work control the compliment of the rationales but with different motivations. Specifically, our work starts with a different motivation by investigating the degeneration problem of cooperative rationalization approaches. We continued the theoretic analysis to alleviate degeneration and arrived at our margin-based objective for practical considerations. In comparison, \citet{carton2018extractive} work on a regression objective, therefore the degeneration problem is not severe as discussed in this paper.

% --------------------------------------------------------------------------------------

\section{Conclusion}
We proposed a novel framework for improving the predictive accuracy and comprehensiveness of the selective rationalization methods.  This framework (1) addresses the degeneration problem in previous cooperative frameworks by regularizing the unselected words via a three-player game; and (2) augments the conventional generator with introspection, which can better maintain the performance for down-stream tasks.  Experiments with both automatic evaluation and subjective studies confirm the advantage of our proposed framework.

\section*{Acknowledgments}
We thank the anonymous reviewers for their helpful comments.
We thank Haoyu Wang and Ming Tan from IBM Watson for the help on human study.
We also thank Yujie Qian, Yujia Bao and Jiang Guo from the MIT NLP group for insightful discussions on the problem of degeneration.

\bibliography{emnlp-ijcnlp-2019}

\begin{thebibliography}{31}
\expandafter\ifx\csname natexlab\endcsname\relax\def\natexlab#1{#1}\fi

\bibitem[{Alvarez-Melis and Jaakkola(2018)}]{alvarez2018towards}
David Alvarez-Melis and Tommi~S Jaakkola. 2018.
\newblock Towards robust interpretability with self-explaining neural networks.
\newblock \emph{arXiv preprint arXiv:1806.07538}.

\bibitem[{Andreas et~al.(2016{\natexlab{a}})Andreas, Rohrbach, Darrell, and
  Klein}]{andreas2016learning}
Jacob Andreas, Marcus Rohrbach, Trevor Darrell, and Dan Klein.
  2016{\natexlab{a}}.
\newblock Learning to compose neural networks for question answering.
\newblock In \emph{Proceedings of NAACL-HLT}, pages 1545--1554.

\bibitem[{Andreas et~al.(2016{\natexlab{b}})Andreas, Rohrbach, Darrell, and
  Klein}]{andreas2016neural}
Jacob Andreas, Marcus Rohrbach, Trevor Darrell, and Dan Klein.
  2016{\natexlab{b}}.
\newblock Neural module networks.
\newblock In \emph{Proceedings of the IEEE Conference on Computer Vision and
  Pattern Recognition}, pages 39--48.

\bibitem[{Arumae and Liu(2019)}]{arumae2019guiding}
Kristjan Arumae and Fei Liu. 2019.
\newblock Guiding extractive summarization with question-answering rewards.
\newblock In \emph{Proceedings of the 2019 Conference of the North American
  Chapter of the Association for Computational Linguistics: Human Language
  Technologies, Volume 1 (Long and Short Papers)}, pages 2566--2577.

\bibitem[{Bastings et~al.(2019)Bastings, Aziz, and
  Titov}]{bastings2019interpretable}
Joost Bastings, Wilker Aziz, and Ivan Titov. 2019.
\newblock Interpretable neural predictions with differentiable binary
  variables.
\newblock \emph{arXiv preprint arXiv:1905.08160}.

\bibitem[{Busoniu et~al.(2006)Busoniu, Babuska, and
  De~Schutter}]{busoniu2006multi}
Lucian Busoniu, Robert Babuska, and Bart De~Schutter. 2006.
\newblock Multi-agent reinforcement learning: A survey.
\newblock In \emph{2006 9th International Conference on Control, Automation,
  Robotics and Vision}, pages 1--6. IEEE.

\bibitem[{Carton et~al.(2018)Carton, Mei, and Resnick}]{carton2018extractive}
Samuel Carton, Qiaozhu Mei, and Paul Resnick. 2018.
\newblock Extractive adversarial networks: High-recall explanations for
  identifying personal attacks in social media posts.
\newblock In \emph{Proceedings of the 2018 Conference on Empirical Methods in
  Natural Language Processing}, pages 3497--3507.

\bibitem[{Chen et~al.(2018{\natexlab{a}})Chen, Song, Wainwright, and
  Jordan}]{chen2018learning}
Jianbo Chen, Le~Song, Martin Wainwright, and Michael Jordan.
  2018{\natexlab{a}}.
\newblock Learning to explain: An information-theoretic perspective on model
  interpretation.
\newblock In \emph{International Conference on Machine Learning}, pages
  882--891.

\bibitem[{Chen et~al.(2018{\natexlab{b}})Chen, Song, Wainwright, and
  Jordan}]{chen2018shapley}
Jianbo Chen, Le~Song, Martin~J Wainwright, and Michael~I Jordan.
  2018{\natexlab{b}}.
\newblock {L-Shapley and C-Shapley}: Efficient model interpretation for
  structured data.
\newblock \emph{arXiv preprint arXiv:1808.02610}.

\bibitem[{Goodfellow et~al.(2014)Goodfellow, Pouget-Abadie, Mirza, Xu,
  Warde-Farley, Ozair, Courville, and Bengio}]{goodfellow2014generative}
Ian Goodfellow, Jean Pouget-Abadie, Mehdi Mirza, Bing Xu, David Warde-Farley,
  Sherjil Ozair, Aaron Courville, and Yoshua Bengio. 2014.
\newblock Generative adversarial nets.
\newblock In \emph{Advances in neural information processing systems}, pages
  2672--2680.

\bibitem[{Hendrickx et~al.(2009)Hendrickx, Kim, Kozareva, Nakov,
  {\'O}~S{\'e}aghdha, Pad{\'o}, Pennacchiotti, Romano, and
  Szpakowicz}]{hendrickx2009semeval}
Iris Hendrickx, Su~Nam Kim, Zornitsa Kozareva, Preslav Nakov, Diarmuid
  {\'O}~S{\'e}aghdha, Sebastian Pad{\'o}, Marco Pennacchiotti, Lorenza Romano,
  and Stan Szpakowicz. 2009.
\newblock Semeval-2010 task 8: Multi-way classification of semantic relations
  between pairs of nominals.
\newblock In \emph{Proceedings of the Workshop on Semantic Evaluations: Recent
  Achievements and Future Directions}, pages 94--99. Association for
  Computational Linguistics.

\bibitem[{Johnson et~al.(2017)Johnson, Hariharan, van~der Maaten, Hoffman,
  Fei-Fei, Lawrence~Zitnick, and Girshick}]{johnson2017inferring}
Justin Johnson, Bharath Hariharan, Laurens van~der Maaten, Judy Hoffman,
  Li~Fei-Fei, C~Lawrence~Zitnick, and Ross Girshick. 2017.
\newblock Inferring and executing programs for visual reasoning.
\newblock In \emph{Proceedings of the IEEE Conference on Computer Vision and
  Pattern Recognition}, pages 2989--2998.

\bibitem[{Kononenko et~al.(2010)}]{kononenko2010efficient}
Igor Kononenko et~al. 2010.
\newblock An efficient explanation of individual classifications using game
  theory.
\newblock \emph{Journal of Machine Learning Research}, 11(Jan):1--18.

\bibitem[{Lee et~al.(2018)Lee, Alvarez-Melis, and Jaakkola}]{lee2018game}
Guang-He Lee, David Alvarez-Melis, and Tommi~S Jaakkola. 2018.
\newblock Game-theoretic interpretability for temporal modeling.
\newblock \emph{arXiv preprint arXiv:1807.00130}.

\bibitem[{Lei et~al.(2016)Lei, Barzilay, and Jaakkola}]{lei2016rationalizing}
Tao Lei, Regina Barzilay, and Tommi Jaakkola. 2016.
\newblock Rationalizing neural predictions.
\newblock In \emph{Proceedings of the 2016 Conference on Empirical Methods in
  Natural Language Processing}, pages 107--117.

\bibitem[{Li et~al.(2016{\natexlab{a}})Li, Chen, Hovy, and
  Jurafsky}]{li2016visualizing}
Jiwei Li, Xinlei Chen, Eduard Hovy, and Dan Jurafsky. 2016{\natexlab{a}}.
\newblock Visualizing and understanding neural models in {NLP}.
\newblock In \emph{Proceedings of the 2016 Conference of the North American
  Chapter of the Association for Computational Linguistics: Human Language
  Technologies}, pages 681--691.

\bibitem[{Li et~al.(2016{\natexlab{b}})Li, Monroe, and
  Jurafsky}]{li2016understanding}
Jiwei Li, Will Monroe, and Dan Jurafsky. 2016{\natexlab{b}}.
\newblock Understanding neural networks through representation erasure.
\newblock \emph{arXiv preprint arXiv:1612.08220}.

\bibitem[{Lundberg and Lee(2017)}]{lundberg2017unified}
Scott~M Lundberg and Su-In Lee. 2017.
\newblock A unified approach to interpreting model predictions.
\newblock In \emph{Advances in Neural Information Processing Systems}, pages
  4765--4774.

\bibitem[{McAuley et~al.(2012)McAuley, Leskovec, and
  Jurafsky}]{mcauley2012learning}
Julian McAuley, Jure Leskovec, and Dan Jurafsky. 2012.
\newblock Learning attitudes and attributes from multi-aspect reviews.
\newblock In \emph{2012 IEEE 12th International Conference on Data Mining},
  pages 1020--1025. IEEE.

\bibitem[{Nguyen and Grishman(2015)}]{nguyen2015combining}
Thien~Huu Nguyen and Ralph Grishman. 2015.
\newblock Combining neural networks and log-linear models to improve relation
  extraction.
\newblock \emph{arXiv preprint arXiv:1511.05926}.

\bibitem[{Ribeiro et~al.(2016)Ribeiro, Singh, and Guestrin}]{ribeiro2016should}
Marco~Tulio Ribeiro, Sameer Singh, and Carlos Guestrin. 2016.
\newblock Why should {I} trust you?: Explaining the predictions of any
  classifier.
\newblock In \emph{Proceedings of the 22nd ACM SIGKDD international conference
  on knowledge discovery and data mining}, pages 1135--1144. ACM.

\bibitem[{Silver et~al.(2017)Silver, Hubert, Schrittwieser, Antonoglou, Lai,
  Guez, Lanctot, Sifre, Kumaran, Graepel et~al.}]{silver2017mastering}
David Silver, Thomas Hubert, Julian Schrittwieser, Ioannis Antonoglou, Matthew
  Lai, Arthur Guez, Marc Lanctot, Laurent Sifre, Dharshan Kumaran, Thore
  Graepel, et~al. 2017.
\newblock Mastering chess and shogi by self-play with a general reinforcement
  learning algorithm.
\newblock \emph{arXiv preprint arXiv:1712.01815}.

\bibitem[{Simonyan et~al.(2013)Simonyan, Vedaldi, and
  Zisserman}]{simonyan2013deep}
Karen Simonyan, Andrea Vedaldi, and Andrew Zisserman. 2013.
\newblock Deep inside convolutional networks: {Visualising} image
  classification models and saliency maps.
\newblock \emph{arXiv preprint arXiv:1312.6034}.

\bibitem[{Srivastava et~al.(2014)Srivastava, Hinton, Krizhevsky, Sutskever, and
  Salakhutdinov}]{srivastava2014dropout}
Nitish Srivastava, Geoffrey Hinton, Alex Krizhevsky, Ilya Sutskever, and Ruslan
  Salakhutdinov. 2014.
\newblock Dropout: a simple way to prevent neural networks from overfitting.
\newblock \emph{The Journal of Machine Learning Research}, 15(1):1929--1958.

\bibitem[{Sundararajan et~al.(2017)Sundararajan, Taly, and
  Yan}]{sundararajan2017axiomatic}
Mukund Sundararajan, Ankur Taly, and Qiqi Yan. 2017.
\newblock Axiomatic attribution for deep networks.
\newblock In \emph{Proceedings of the 34th International Conference on Machine
  Learning-Volume 70}, pages 3319--3328. JMLR. org.

\bibitem[{Williams(1992)}]{Williams1992}
Ronald~J. Williams. 1992.
\newblock Simple statistical gradient-following algorithms for connectionist
  reinforcement learning.
\newblock \emph{Machine Learning}.

\bibitem[{Yala et~al.(2019)Yala, Lehman, Schuster, Portnoi, and
  Barzilay}]{yala2019deep}
Adam Yala, Constance Lehman, Tal Schuster, Tally Portnoi, and Regina Barzilay.
  2019.
\newblock A deep learning mammography-based model for improved breast cancer
  risk prediction.
\newblock \emph{Radiology}, page 182716.

\bibitem[{Yu et~al.(2018)Yu, Chang, and Jaakkola}]{yu2018learning}
Mo~Yu, Shiyu Chang, and Tommi~S Jaakkola. 2018.
\newblock Learning corresponded rationales for text matching.

\bibitem[{Zaidan et~al.(2007)Zaidan, Eisner, and Piatko}]{zaidan2007using}
Omar Zaidan, Jason Eisner, and Christine Piatko. 2007.
\newblock Using “annotator rationales” to improve machine learning for text
  categorization.
\newblock In \emph{Human language technologies 2007: The conference of the
  North American chapter of the association for computational linguistics;
  proceedings of the main conference}, pages 260--267.

\bibitem[{Zhang et~al.(2017)Zhang, Barzilay, and Jaakkola}]{zhang2017aspect}
Yuan Zhang, Regina Barzilay, and Tommi Jaakkola. 2017.
\newblock Aspect-augmented adversarial networks for domain adaptation.
\newblock \emph{Transactions of the Association for Computational Linguistics},
  5:515--528.

\bibitem[{Zhao et~al.(2017)Zhao, Yue, Katabi, Jaakkola, and
  Bianchi}]{zhao2017learning}
Mingmin Zhao, Shichao Yue, Dina Katabi, Tommi~S Jaakkola, and Matt~T Bianchi.
  2017.
\newblock Learning sleep stages from radio signals: A conditional adversarial
  architecture.
\newblock In \emph{Proceedings of the 34th International Conference on Machine
  Learning-Volume 70}, pages 4100--4109. JMLR. org.

\end{thebibliography}
\bibliographystyle{acl_natbib}

\appendix
\clearpage

\section{Proof of Theorem \ref{thm:main}}
\label{app:theorem}

This appendix provides the proof of theorem \ref{thm:main}.  First, we will need the following lemma.
\begin{lemma}
If the predictors have sufficient representation power, we have
\begin{equation}
    \small
    \mathcal{L}_p = -H(Y|\bm R), \quad \mathcal{L}_c=-H(Y|\bm R^c).
\end{equation}
\label{lem:entropy}
\end{lemma}
The proof is trivial, noticing cross entropy is upper bounded by entropy.

The following lemmas show that there is a correspondence between the rationale properties in Eqs.~\eqref{eq:suff} and loss terms in Eq.~\eqref{eq:loss_gen}.
\begin{lemma}
A rationalization scheme \e{\bm z (\bm X)} that satisfies Eq.~\eqref{eq:suff} is the global minimizer of \e{\mathcal{L}_p} as defined in Eq.~\eqref{eq:loss_pred}.
\label{lem:Lp}
\end{lemma}
\begin{proof}
Notice that
\begin{equation}
    \small
    \mathcal{L}_p = -H(Y|\bm R) = -H(Y| \bm r(\bm X)) \geq -H(Y | \bm X).
\end{equation}
The first equality is given by Lemma \ref{lem:entropy}; the second equality is given by Eq.~\eqref{eq:notation}. For the inequality, the equality holds if and only if
\begin{equation}
    \small
    p_Y(\cdot | \bm r(\bm X)) = p_Y(\cdot | \bm X),
\end{equation}
which is Eq.~\eqref{eq:suff}.
\end{proof}

\begin{lemma}
A rationalization scheme \e{\bm z (\bm X)} that satisfies Eq.~\eqref{eq:nondegen} is the global minimizer of \e{\mathcal{L}_g} as defined in Eq.~\eqref{eq:Lg}.
\label{lem:Lg}
\end{lemma}
\begin{proof}
According to Lemma \ref{lem:entropy}, \e{\mathcal{L}_g} can be rewritten as
\begin{equation}
    \small
    \mathcal{L}_g = \max\{H(Y|\bm R)-H(Y|\bm R^c)+h, 0\},
\end{equation}
which equals 0 if and only if Eq.~\eqref{eq:nondegen} holds.
\end{proof}
\begin{lemma}
A rationalization scheme \e{\bm z (\bm X)} that satisfies Eq.~\eqref{eq:compact} is the global minimizer of \e{\mathcal{L}_s} and \e{\mathcal{L}_c} as defined in Eq.~\eqref{eq:Ls_Lc}.
\label{lem:Ls_Lc}
\end{lemma}
\begin{proof}
The proof is obvious. \e{\mathcal{L}_s} and \e{\mathcal{L}_c} is 0 if and only if Eq.~\eqref{eq:compact} holds.
\end{proof}

Combining Lemmas \ref{lem:Lp} to \ref{lem:Ls_Lc} completes the proof of Theorem \ref{thm:main}.

%----------------------------------------------
\section{Experimental Setup of Examples in Table \ref{tab:example} and Degeneration Cases of \cite{lei2016rationalizing}}
\label{app:taos_degeneration}

This section provides the details to obtain the results in Table \ref{tab:example} in the introduction section, where the method of \cite{lei2016rationalizing} generates degenerated rationales.

The method of~\cite{lei2016rationalizing} works well in many applications. However, as discussed in Section~\ref{sec:intro} and \ref{ssec:axiomatic}, all the cooperative rationalization approaches may suffer from the problem of degeneration.
In this section, we design an experiment to confirm the existence of the problem in the original~\cite{lei2016rationalizing} model. 
We use the same single-beer review constructed from \cite{mcauley2012learning}, as will be described in Appendix~\ref{app:data_construction}. 
Instead of constructing a balanced binary classification task, we set the samples with scores higher than 0.5 as positive examples.
On such a task, the prediction model with full inputs achieves 82.3\% accuracy on the development set.

During the training of \cite{lei2016rationalizing}, we stipulate that the generated rationales are very concise: we punish it when the rationales have more than 3 pieces or more than 20\% of the words are generated (both with hinge losses).
From the results, we can see that \citet{lei2016rationalizing} tends to predict color words, like \emph{dark-brown}, \emph{yellow}, as rationales.
This is a clue of degeneration, since most of the appearance reviews start with describing colors. Therefore a degenerated generator can learn to split the vocabulary of colors, and communicate with the predictor by using some of the colors for the positive label and some others for the negative label.
Such a learned generator also fails to generalize well, given the significant performance decrease (76.4\% v.s. 82.3\%).
By comparison, our method with three-player game could achieve both higher accuracy and more meaningful rationales.

\begin{table*}[t]
\small
\centering
\begin{tabular}{lccccc}
\toprule
\multicolumn{1}{c}{\multirow{1}{*}{\bf Datasets}} & \multirow{1}{*}{\bf \# Classes} & \bf Vocab Size                                   & \multicolumn{1}{c}{\bf \# Train} & \multicolumn{1}{c}{\bf \# Dev} & \multicolumn{1}{c}{\bf \# Annotation/Test} \\
\midrule
\multirow{1}{*}{Multi-aspect sentiment classification}    &2 & 110,985 & 80,000 & 10,000    & 994            \\
\multirow{1}{*}{Single-aspect sentiment classification}   &2  & 12,043 &  12,000 & 1,362    & 1,695          \\
\multirow{1}{*}{Relation classification}    & 19 & 23,446 & 7,000    & 1,000      & 2,717\\
\bottomrule
\end{tabular}
\caption{\small {Statistics of the datasets used in this paper. }}
\label{tab:dataset_stats}
\end{table*}

\begin{table}[h]
\small
\centering
\begin{tabular}{ll}
\toprule
\multicolumn{2}{p{7cm}}{\textbf{Original Text} (positive): \textit{dark-brown/black color with a \textcolor{red}{\textit{huge tan head}} that gradually collapses , leaving \textcolor{red}{\textit{thick lacing}} .}} \\
\midrule
\textbf{Rationale from \cite{lei2016rationalizing}} (Acc: 76.4\%):\\ \multicolumn{2}{p{7cm}}{[\textit{``dark-brown/black color''}]} \\
\textbf{Rationale from our method} (Acc: 80.4\%): \\
\multicolumn{2}{p{7cm}}{[\textit{``huge tan'', ``thick lacing''}]}\\
\midrule
\multicolumn{2}{p{7cm}}{\textbf{Original Text} (negative): \textit{really \textcolor{red}{\textit{cloudy}} , \textcolor{red}{\textit{lots of sediment}} , washed out yellow color . looks \textcolor{red}{\textit{pretty gross}} , actually , like swamp water . \textcolor{red}{\textit{no head}} , \textcolor{red}{\textit{no lacing}} .}} \\
\midrule
\textbf{Rationale from \cite{lei2016rationalizing}} (Acc: 76.4\%):\\ \multicolumn{2}{p{7cm}}{[\textit{``really cloudy lots'', ``yellow'', ``no'', ``no''}]} \\
\textbf{Rationale from our method} (Acc: 80.4\%): \\
\multicolumn{2}{p{7cm}}{[\textit{``cloudy'', ``lots'', ``pretty gross'', ``no lacing''}]}\\
\bottomrule
\end{tabular}
\caption{\small An example showing rationales extracted by different models, where \cite{lei2016rationalizing} gives degenerated result.}
\label{tab:example_bak}
\end{table}

\section{Data Construction of the Single-Aspect Beer Reviews}
\label{app:data_construction}

This section describes how we construct the single-aspect review task from the multi-aspect beer review dataset~\cite{mcauley2012learning}.

In many multi-aspect beer reviews, we can see clear patterns indicating the aspect of the following sentences.
For example, the sentences starting with ``\emph{appearance:}'' or  ``\emph{a:}'' are likely to be a review on the appearance aspect; and the sentences starting with ``\emph{smell:}'' or  ``\emph{nose:}'' are likely to be about the aroma aspect.

We then extract all the ``\emph{X:}'' patterns, and count the frequencies of such patterns, where each \emph{X} is a word.
The patterns ``\emph{X:}'' with a higher than 400 frequency are kept as \textbf{anchor patterns}. The sentences between two anchor patterns ``\emph{X$_1$: $\cdots$ X$_2$}'' are very likely the review regarding the aspect of \emph{X$_1$}.
Finally, we extract such review sentences after ``\emph{appearance:}'' or  ``\emph{a:}'' and before the immediate subsequent anchor patterns as the single-aspect review for the appearance aspect. Each of such instances is regarded as a new single-aspect review.  The score of the appearance aspect of the original multi-aspect review is regarded as the score of this new review.

With such an automatically constructed dataset, we form our balanced single-review binary classification tasks (see Section \ref{ssec:datasets} and Appendix \ref{app:data_statistics}), on which our base predictor model (with all the words as inputs) performs an 87.1\% on the development set. This is as high as the number we achieved on the multi-aspect task regarding the same aspect (87.6\%). This result indicates that the noise introduced by our data construction method is insignificant.

\section{Data Statistics}
\label{app:data_statistics}

Table \ref{tab:dataset_stats} summarizes the statistics of the three datasets used in the experiments. The single-aspect sentiment classification and the relation classification have randomly held-out development sets from the original training sets.

\section{Experiment Designs for Human Study}
\label{app:guidelines}

This section explains how we designed the human study.

The goal is to evaluate the unpredictable rates of the input texts after the rationales are removed. To this end, we mask the original texts with the rationales generated by \cite{lei2016rationalizing} and our method. Each rationale word is masked with the symbol `\emph{*}'.
The masked texts from different methods are mixed and shuffled so the evaluators cannot know from which systems an input was generated.

We have two human evaluators who are not the authors of the paper.
During evaluation, an evaluator is presented with one masked text and asked to try her/his best to predict the sentiment label of it.
If a rationalizing method successfully includes all informative pieces in the rationale, subjects should have around 50\% of accuracy in guessing the label.

After the evaluator provides a sentiment label, the subjects are asked to answer the second question about whether the provided text spans are sufficient for them to predict the sentiment.
If they believe there are no enough clues and their sentiment classification is based on a random guess, they are instructed to input a \emph{UNK} label as the answer to the second question.

The reason we ask the evaluators to provide predicted labels first is based on the following idea: if the task is directly annotating whether the masked texts are unpredictable, the annotators will tend to label more \emph{UNK} labels to save time. Therefore the ratios of \emph{UNK} labels will be biased.
Our experimental design alleviates this problem since the evaluators are always required to try the best to guess the labels first. Therefore they will spend more time thinking about the possible labels, instead of immediately putting a \emph{UNK} label.

On a small subset of 50 examples, the inter-annotator agreement is 76\% on the \emph{UNK} labels.

%%%%%%%%%%%%%%%%%%%%%%%%%%%%%%%%%%%%%%%%%%%%%%%%%%%%%%%%%%%%%%%
\section{Additional Experiments on AskUbuntu}
\label{app:askubuntu}
\paragraph{Setting}

Following the suggestion from the reviews, we evaluate the proposed method on the question retrieval task on AskUbuntu \cite{lei2016rationalizing}.  AskUbuntu is a non-factoid question retrieval benchmark.  The goal is to retrieve the most relevant questions from an input question.  We use the same data split provided by~\cite{lei2016rationalizing}.\footnote{\url{https://github.com/taolei87/askubuntu}.}

Specifically, each question consists of two parts, the \emph{question title} and the \emph{question body}.   The former summarizes a problem from using Ubuntu, while the latter contains the detailed descriptions.  In our experiments, we follow the same setting from \cite{lei2016rationalizing} by only using the question bodies. Different from their work, we do not 
pre-train an encoder by predicting a question title using the corresponding question body.   
This is because, the question title can be considered as the rationale of its question body, which might result in potential information leaks to our main rationalization task.  

%%%%%%%%%%%%%%%%%%%%%%%%%%%%%%%%%%%%%%%%%%%%%%%%%%%%%%%%%%%%%%%
\paragraph{Method}
We formulate the problem of question retrieval as the pairwise classification task. Given two questions (\emph{i.e.}, a query and a candidate question),  we aim to classify them as a positive label if they are relevant and vice-versa.  We consider the same generator architecture as used in Section~\ref{sec:exp} in a siamese setting to extract rationales from questions.  The predictor and the complement predictor make the prediction based on the pairwise selected spans.  We believe it is the most straightforward way to adapt the proposed framework to the AskUbuntu task.  There could be sophisticated and task-specific rationalization approaches to improve the performance on AskUbuntu (\emph{e.g.}, using a ranking model instead of a classification model).  However, newly design of introspective modules are also required.  We leave these investigations to future works.

%%%%%%%%%%%%%%%%%%%%%%%%%%%%%%%%%%%%%%%%%%%%%%%%%%%%%%%%%%%%%%%
\paragraph{Implementation Details}
We consider the following three-step training strategy: 1) pre-train a classifier with the full text;  2) fix the pre-trained classifier, which is used for both the predictor and the complement predictor in the three-player game approach, and pre-train the rationale generators; and 3) fine-tune all modules end-to-end.  This pipeline significantly stabilizes the training and provides better performances.\footnote{One potential reason that the three-step training strategy performs much better than end-to-end training from scratch is that we sample rationales according to the policy $\pi(\cdot)$ during training but take the action with the highest probability during the inference.  During the first a few epochs of training, rationale generator almost extracts words at any positions with a probability lower than 0.5. Rationale words are still able to be sampled during training.  However, during inference, there are no rationale words selected unless a probability of selection is greater than 0.5.  Thus, the MAP on the development set is unchanged at the beginning stage of the training.   In other words,  there is a risk that the predictor already overfits but we cannot perform early-stopping of the training.}   We use the same word embeddings as released by \cite{lei2016rationalizing}.

%%%%%%%%%%%%%%%%%%%%%%%%%%%%%%%%%%%%%%%%%%%%%%%%%%%%%%%%%%%%%%%
\paragraph{Results}
Table \ref{tab:askubuntu_map} summarizes the results.  We observe similar patterns as in previous datasets.  The original model from \cite{lei2016rationalizing} fails to maintain the performance compared to the model trained with full texts.  Adding the proposed minimax game helps both the \cite{lei2016rationalizing} and the introspection model to generate more informative texts as the rationales, which improves the MAP of the prediction while lowering the complement MAP.

Compared to the other tasks, the complement MAPs on AskUbuntu are relatively large.   One reason is that the reported results rely significantly on the three-step training strategy.   The best MAP on the development set often occurs after a few epochs of end-to-end training (the third step of our training procedure), 
which may results in premature training of the generators due to early stop.
Another important reason is that there are a larger number of informative words in the questions, which makes it challenging for the generators to include all the useful information.

\begin{table}[t]
\centering
\small
\begin{tabular}{lcccc}
\toprule
% \midrule
\multirow{1}{*}{\bf Model} & \bf Highlight Percentage  & \bf MAP & \bf MAP$^{c}$ \\
\midrule
All & 100\% & 51.55 & 38.97  \\
\midrule
Lei2016 &  20\%& 43.64 & 47.84 \\
\multirow{1}{*}{ +minimax}& 20\% & 48.58 & 46.13 \\
\midrule
Intros &  20\%&  45.08 & 49.27 \\
\multirow{1}{*}{ +minimax}& 20\% & 48.55 & 48.37 \\
\bottomrule
\end{tabular}
\caption{\small{Testing MAP on the AskUbuntu dataset.  MAP$^c$ refers to the MAP score of the complement predictor. The desired rationalization method will have high MAP and low MAP$^c$.}}
\label{tab:askubuntu_map}
\end{table}

\end{document}